%% file: main.tex
\documentclass{article}
\usepackage{amsmath}

\usepackage[preprint]{corl_2026} 

\usepackage[utf8]{inputenc} 
\usepackage[T1]{fontenc}    
\usepackage{hyperref}       
\usepackage[capitalise]{cleveref}
\usepackage{algorithm}
\usepackage{algpseudocode}
\usepackage{amsmath}
\usepackage{url}            
\usepackage{booktabs}       
\usepackage{amsfonts}       
\usepackage{nicefrac}       
\usepackage{microtype}      
\usepackage{xcolor}         

\usepackage{graphicx} 
\usepackage{booktabs}    
\usepackage{multirow}    
\usepackage{makecell} 
\usepackage{caption}
\usepackage{wrapfig}
\usepackage{enumitem}
\usepackage{tcolorbox}
\usepackage{listings}
\definecolor{systembg}{RGB}{232, 245, 233}
\definecolor{userbg}{RGB}{227, 242, 253}

\newcommand{\ours}{Mem-World }

\title{Mem-World: Memory-Augmented Action-Conditioned World Models for Persistent Robot Manipulation}

\makeatletter
\renewcommand{\@fnsymbol}[1]{}
\makeatother

\author{Zirui Zheng$^{1,2}$, Jiaqian Yu$^2$, Xiongfeng Peng$^2$, Jun Shi$^2$, Mingyi Li$^1$,\\ \textbf{ Chao Zhang$^2$, Weiming Li$^2$, Dong Wang$^1$, Huchuan Lu$^1$, Xu Jia$^{1,*}$}
\thanks{$^{1}$Dalian University of Technology, $^{2}$Samsung R\&D Institute China-Beijing (SRCB), China}
\thanks{$^{*}$Corresponding author}
}
\begin{document}



\maketitle

\setlength{\textfloatsep}{2pt plus 1pt minus 1pt}
\setlength{\intextsep}{2pt plus 1pt minus 1pt}
\setlength{\floatsep}{2pt plus 1pt minus 1pt}

\input{sec/0_abstract}   

\input{sec/1_intro}

\input{sec/2_related_work}

\input{sec/3_method}

\input{sec/4_experiments}
\input{sec/5_conclusion}

\input{sec/6_limitation}

\clearpage
\bibliography{main}  

\end{document}

%% file: sec/0_abstract.tex
\begin{abstract}
Action-conditioned world models have emerged as a promising paradigm for robot learning, offering a scalable alternative to costly real-world experimentation by generating action-consistent video rollouts. 
However, persistent world modeling remains challenging in manipulation: frequent end-effector occlusions and rapid wrist-camera motion make the current observation insufficient for predicting future views, causing models to forget or hallucinate scene details seen in earlier frames. Existing memory retrieval strategies often fail to identify informative history in dynamic manipulation scenarios.
To address this limitation, we propose Mem-World, a memory-augmented multi-view action-conditioned world model. At its core, we present W-VMem, a 4D wrist-view-centered surfel-indexed memory that anchors historical observations to temporally evolving surface elements. By explicitly modeling when and where scene elements are observed, W-VMem enables geometry-aware retrieval of relevant history frames conditioned on future actions. During generation, relevant history frames are selected via surfel-based rendering and scoring, providing informative and non-redundant context for prediction. 
Extensive experiments show that Mem-World generates persistent rollouts in complex manipulation scenarios, 
enables more reliable policy evaluation than Ctrl-World, improving the Pearson correlation with real-world performance by 14.5\%, and supports effective policy improvement through synthetic data generation, increasing success rates from 58\% to 72\% on long-horizon tasks.

\end{abstract}
\keywords{World Models, Robot Manipulation, Memory Mechanism} 


%% file: sec/1_intro.tex
\section{Introduction}
Action-conditioned world models represent a rapidly advancing direction in robot learning~\citep{guo2025ctrl-world,jiang2025enerverse, fu2025robomaster, li2024ewpe, zhu2025irasim, wang2026interactivews, ali2025cosmos, assran2025vjepa2}, offering a scalable alternative to costly real-world experimentation. Built on pretrained video diffusion backbones~\citep{blattmann2023svd, zheng2024opensora, wan2025wan} and fine-tuned with action conditioning on large-scale robotics datasets~\citep{khazatsky2024droid, ebert2021bridge, o2024openx}, these models generate action-consistent video trajectories that can serve as virtual environments for policy development and evaluation. 

In robot manipulation, such world models are often built in a multi-view setting, combining external third-person cameras with a wrist-mounted camera to capture both global scene context and fine-grained hand-object interactions. This multi-view formulation is important for manipulation, but it also exposes a key challenge: the wrist view is both highly informative and highly unstable. Since the wrist camera moves with the end effector, it undergoes rapid egocentric motion and is frequently occluded by the end-effector or manipulated objects. As a result, the current observation often contains only partial evidence about the scene state, causing autoregressive world models to forget or hallucinate objects that were visible in earlier frames, particularly severe in long-horizon rollouts. 

Existing approaches only partially address this issue. Interactive World Simulator~\citep{wang2026interactivews} mitigates limited temporal context by adjusting the conditioning window size, predicting future observations from the current observation together with adjacent $N$ historical frames; however, simply enlarging the context window does not essentially solve the problem. Ctrl-World~\citep{guo2025ctrl-world} addresses history-frame retrieval by sampling $N$ conditioning frames from the entire history at a fixed stride and using joint-pose similarity to emphasize historical frames with similar joint configurations during future-observation prediction. However, joint-pose similarity is only an indirect proxy for contextual relevance. 
It fails to capture visibility constraints introduced by end-effector occlusions and cannot prioritize viewpoints that provide richer scene coverage, such as elevated wrist-camera observations. 
Similar memory retrieval mechanisms have been explored in navigation-oriented world models, often using camera field-of-view (FOV) overlap~\citep{yu2025context, xiao2026worldmem} or static 3D anchors~\citep{li2025vmem} to retrieve the relevant historical frames. However, manipulation violates the underlying assumptions: the scene is partially dynamic and the robot arm itself causes structured occlusions. 

To address the limitations, we propose Mem-World, a memory-augmented multi-view action-conditioned world model for persistent robot manipulation. At its core, Mem-World introduces the W-VMem, a 4D wrist-view-centered surfel-indexed view memory module inspired by VMem~\citep{li2025vmem}, which anchors previous wrist-view observations to the 4D surface elements they correspond to. Specifically, we first construct a global W-VMem from the initial multi-view frames using an off-the-shelf point map estimator~\citep{lin2025depthany3}. During autoregressive generation of future observations, W-VMem is updated using only the wrist-view observations to capture the association between 4D surfels and the dynamic wrist view. Each surfel in W-VMem maintains its creation timestep, merged timestep, orientation, depth, and a binary flag indicating whether it belongs to a manipulated object. Before generating observations at subsequent timesteps, the memory reading module retrieves relevant historical frames by rendering the surfels at each timestep from future wrist-camera viewpoints and projecting them onto an image grid. Based on the rendered surfel maps and the stored surfel attributes, the module computes a surfel score for each historical timestep and selects observations corresponding to the top-$K$ timesteps as conditioning inputs. By grounding future predictions in these geometrically relevant contexts, Mem-World preserves persistent scene information over long-horizon rollouts.

Extensive experiments show that Mem-World generates more temporally consistent robot manipulation rollouts than prior methods. To specifically evaluate memory persistence, we curate replay trajectories with frequent end-effector occlusions and complex wrist-camera motions. Mem-World outperforms prior world models across frame-level and object-level consistency metrics. It also provides more reliable policy evaluation, improving the Pearson correlation with real-world performance by 14.5\%, and improves long-horizon policy performance with generated data, increasing the average success rate from 58\% to 72\%. Our contributions are summarized as follows:
\begin{itemize}[leftmargin=8pt]
\item  We propose Mem-World, a memory-augmented multi-view action-conditioned world model for persistent robot manipulation rollouts, improving the reliability of world models for long-horizon policy evaluation and training.
\item We introduce W-VMem, a 4D wrist-view-centered surfel-indexed memory that anchors historical observations to temporally evolving surface elements and enables geometry-aware retrieval under severe occlusions, rapid wrist-camera motion, and dynamic hand-object interactions.
\item We evaluate Mem-World under memory-stressing replay trajectories and long-horizon manipulation tasks, showing more temporally consistent rollouts, stronger correlation with real-world policy performance, and improved policy learning from synthetic data.
\end{itemize}



%% file: sec/2_related_work.tex
\section{Related Work}

\noindent \textbf{Video Prediction Model for Robotic Manipulation.}
Recent video prediction advances have demonstrated substantial promise for robotics, enabling models to serve as zero-shot planners~\citep{chen2025videoplan, li2025novaflow}, direct controllers~\citep{du2023unipi,pai2025mimic,ye2026dreamzero, huang2025ladi}, and data engines~\citep{jang2025dreamgen, team2025gigaworld, wang2026robovip}. Complementary to these applications, action-conditioned world models~\citep{zhu2025irasim, guo2025ctrl-world,  jiang2025enerverse, chandra2025diwa, wang2026interactivews} incorporate low-level actions as conditioning signals, making them particularly suitable for policy evaluation. IRASim~\citep{zhu2025irasim} formulates robot manipulation as a trajectory-to-video generation problem and introduces frame-level action conditioning. Ctrl-World~\citep{guo2025ctrl-world} proposes a multi-view action-conditioned world model that simultaneously generates two third-person views and one wrist view, capturing manipulation details from complementary perspectives. Interactive World Simulator~\citep{wang2026interactivews} proposes a head-view action-conditioned world model built upon a consistency model, enabling long-horizon real-time closed-loop rollouts and policy evaluation. Despite this progress, persistent long-horizon multi-view rollout remains underexplored. To address this gap, we introduce \ours, a multi-view, action-conditioned world model with a 4D, wrist-view-centered, surfel-indexed memory module that retrieves relevant historical frames during rollouts. Extensive experiments show \ours supports reliable policy evaluation and policy improvement in long-horizon tasks by maintaining persistent multi-view world modeling.

\noindent \textbf{Memory Modeling in Long-Horizon Video Generation.}
Maintaining long-horizon consistency is a central challenge in world models; existing efforts address this issue through either explicit 3D reconstruction~\citep{yu2024viewcrafter, cao2025uni3c, ren2025gen3c} or implicit context conditioning~\citep{xiao2026worldmem, yu2025context, li2025vmem, sun2025worldplay}. Explicit methods leverage 3D foundation models to reconstruct scene representations and render condition frames from novel viewpoints, providing strong geometric guidance but relying heavily on reconstruction quality and introducing substantial computational overhead. Implicit methods maintain a memory bank and retrieve spatially relevant history to condition the current generation. Geometry-aware retrieval strategies use camera poses or field-of-view overlap~\citep{xiao2026worldmem, yu2025context} to identify informative context. VMem~\citep{li2025vmem} further proposes a lightweight surfel-indexed memory that associates static 3D surfels with historical observations and retrieves relevant frames by rendering the memory from future camera poses.
Our work differs by making memory retrieval temporally aware and manipulation-centric, using dynamic surface-level associations rather than static camera-pose overlap alone.


\vspace{-4mm}

%% file: sec/3_method.tex
\begin{figure}[t]
    \centering
    \includegraphics[width=0.98\linewidth]{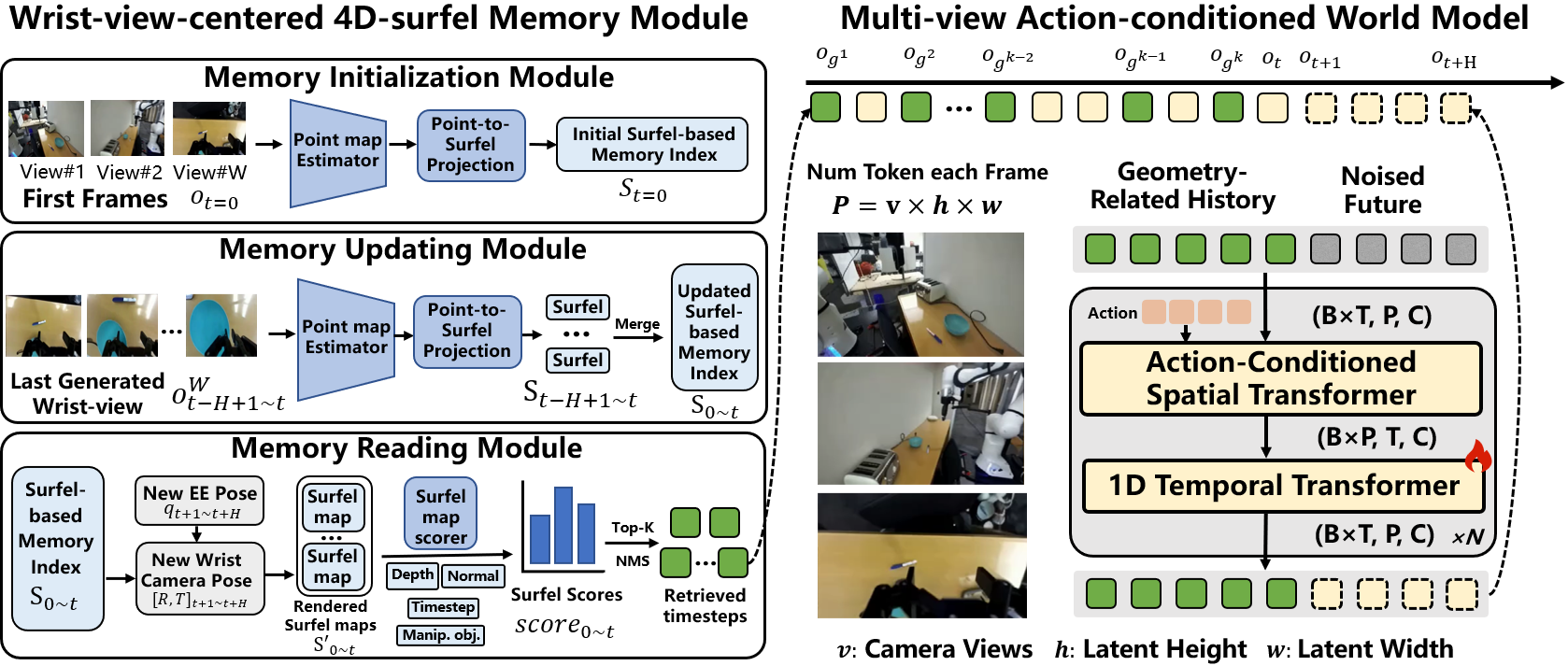}
    \caption{Pipeline of Mem-World. During rollout, a wrist-view 4D surfel memory maintains a geometry-aware representation of past observations, from which relevant history frames are retrieved. Given these retrieved frames and a sequence of future actions, the multi-view action-conditioned world model predicts subsequent observations. The predicted observations are then integrated back into the memory, enabling iterative rollout with an updated memory module.}
    \label{fig:pipeline}
\end{figure}
\section{Method}
Our goal is to build a multi-view action-conditioned world model that can generate persistent future observations. As shown in \cref{fig:pipeline}, our pipeline consists of two modules: (1) wrist-view-centered 4D surfel memory and (2) a multi-view action-conditioned world model. During rollout, the world model generates future observations conditioned on the retrieved history frames and future actions. The memory module then takes the predicted observations as input and updates W-VMem to support memory retrieval at subsequent timesteps.
In this section, we first introduce the preliminaries of the surfel-indexed view memory in Sec. 3.1, and then describe the memory initialization, reading, and updating procedures of W-VMem in Sec. 3.2.
\subsection{Preliminary}
\textbf{Surfel-indexed View Memory.}
For autoregressive video prediction conditioned on camera pose, VMem~\citep{li2025vmem} constructs a Surfel-indexed view memory based on past observations (view), where each surfel $s_k=(p_k,n_k,r_k,\mathcal{I}_k)$ stores its 3D position $p_k$, surface normal $n_k$, radius $r_k$, and the indices $\mathcal{I}_k$ of observations that have observed it. Given a future camera pose, VMem renders all visible surfels and aggregates their associated observation indices. It then retrieves historical observations that cover larger portions of the rendered surfels, using them as context for predicting the future observation. The predicted observations are then converted into surfels and merged into memory. This iterative geometry-aware retrieval and update process improves long-term consistency in static egocentric scenes, but it assumes that camera poses are directly available and does not explicitly distinguish temporally changing scene elements.

\subsection{W-VMem for Persistent Manipulation World Modeling}

A standard action-conditioned world model $W$ predicts future observations $\hat{o}_{t+1:t+H}$ conditioned on the current observation $\hat{o}_t$ and the action chunk $a_{t+1:t+H}$, Mem-World augments the conditions with additional historical observations. By extending Surfel-indexed View Memory to multi-view action-conditioned manipulation world models, W-VMem retrieves a set of relevant history frames
$\mathcal{H}_t=\{o_{g_1},\ldots,o_{g_K}\}$ at rollout timestep $t$,
where $\{g_i\}_{i=1}^{K}$ are selected historical timesteps. The world model then predicts future multi-view observations as:
\begin{equation}
    \hat{o}_{t+1:t+H}
    \sim
    W(\cdot \mid \hat{o}_{t}, \mathcal{H}_t, a_{t+1:t+H}).
\end{equation}
However, such extending presents three main challenges: (1) Given camera settings for robot manipulation task, how to obtain future camera poses from joint-space actions for rendering surfels? (2) How to define surfels for dynamic robot manipulation scenes? (3) How to initialize, update, and read surfels in multi-view robot manipulation scenes?

\textbf{Grounding future wrist views from robot actions. }
In our multi-view setup, the third-person cameras remain fixed throughout an episode, so future pose estimation only applies to the wrist camera. We exploit the rigid mounting between the wrist camera and the end effector, which provides a fixed geometric transformation between them. Given the future actions predicted by the policy model, we first use forward kinematics to obtain the corresponding end-effector poses, and then apply this fixed transformation to derive future wrist-camera poses in the robot base frame. 
For surfel rendering, these camera poses must be expressed in the frame of the point cloud. In practice, we estimate the point cloud using a calibrated third-person camera as the reference frame, and therefore use the known camera extrinsics to transform the wrist-camera poses.

\textbf{Time-aware Surfel Definition for Manipulation. }
VMem assumes a static scene and renders all surfels jointly, which is insufficient for robot manipulation where the robot arm and manipulated objects move over time. W-VMem augments each surfel with temporal and task-relevance attributes:
$s_k = (p_k, n_k, r_k, t_k, m_k)$, 
where $t_k$ denotes the timestep set at which the surfel is created and updated, enabling the representation to capture temporal dynamics. $m_k \in \{0,1\}$ is a binary flag indicating whether the surfel belongs to a manipulated object, allowing us to emphasize task-relevant regions. To obtain $m_k$,  we use a vision-language model~\citep{qwen3technicalreport} to analyze the scene from the text instruction and initial observation, identifying task-relevant object categories. We then apply a segmentation model~\citep{yuan2025roboengine} to segment them in the image and assign the flag to associated surfels.

\textbf{Memory Initialization and Updating.}
For \textit{initialization}, we construct an initial set of surfels from the first frame of the three camera views, which provides a complete observation of the scene. For observations generated at subsequent timesteps, we \textit{update} the surfel representation using only the wrist-view frame. This design is motivated by two considerations. First, the wrist view provides the most manipulation-centric observation; it directly captures the region under active interaction, where object geometry, texture, contact states, and fine-grained changes are more clearly observed. Second, W-VMem is designed to associate surfels with relevant historical wrist-view observations. If third-person observations were used to update the surfels, their global field of view would cause most visible surfels to be repeatedly refreshed and assigned almost all timestamps. As a result, the timestamps of surfels would no longer indicate when they were observed by the moving wrist camera, making it difficult to associate surfels with the corresponding historical wrist-view observations. By restricting updates to wrist-view observations, W-VMem preserves this temporal association and better records which regions have been actively observed over time.
\begin{wrapfigure}{r}{0.68\textwidth}
\vspace{-0pt}
\begin{minipage}{0.68\textwidth}
\begin{algorithm}[H]
\caption{Mem-World Rollout with W-VMem}
\label{alg:ctrl_mem}
\centering
\scalebox{0.88}{%
\begin{minipage}{1.12\linewidth}
\begin{algorithmic}[1]
\Require Initial multi-view observations $o_{t=0}$, action-conditioned world model $W$, top-$K$, action horizon $H$, interaction step $T$
\Ensure Generated multi-view rollout $\mathcal{O}$
\State Initialize W-VMem $\mathcal{S} \leftarrow \textsc{InitSurfels}(o_0)$,  $\mathcal{O} \gets \mathcal{O} \cup \{o_{t=0}\}$
\For{$t=0$ to $T-1$}
    \State Obtain action chunk $\mathbf{a}_{t+1:t+H}$ from policy or replayed trajectory
    \State Estimate camera pose $\bar{\mathbf{c}}_w \leftarrow \textsc{AvgWristPose}(a_{t+1:t+H})$ 
    \State Compute $score_{\tau} \leftarrow \textsc{Score}(\textsc{Render}(\mathcal{S}_{\tau}, \bar{\mathbf{c}}_w))$ for each $\tau$
    \State Retrieve history frames $\mathcal{H}_t \leftarrow \textsc{TopK-NMS}(\{score_{\tau}\}, K)$
    \State Generate $\hat{o}_{t+1:t+H} \leftarrow W(\hat{o}_{t}, \mathcal{H}_t, \mathbf{a}_{t+1:t+H})$
    \State $\mathcal{O} \gets \mathcal{O} \cup \{\hat{o}_{t+1}, \ldots, \hat{o}_{t+H}\}$
    \State Update W-VMem $\mathcal{S} \leftarrow \textsc{UpdateSurfels}(\mathcal{S}, \hat{o}_{t+1:t+H}^{w})$
\EndFor
\end{algorithmic}
\end{minipage}%
}
\end{algorithm}
\end{minipage}
\vspace{-2pt}
\end{wrapfigure}
\textbf{Geometry-aware Memory Retrieval.}
After updating the W-VMem, the world model would receive a new action chunk from the policy model for future observation generation. To provide historical context for this process, W-VMem is read to retrieve the historical observations relevant to future states. Specifically, we first compute the average future wrist camera pose $\bar{\mathbf{c}}_w \in \mathrm{SE}(3)$ based on the action chunk. Then we render the surfels $\mathcal{S}$ at each timestep individually from the $\bar{\mathbf{c}}_w$, instead of rendering the entire surfel set jointly like VMem. 
This design exploits wrist-view-dependent temporal dynamics, as surfels observed at different timesteps correspond to distinct interaction stages and should not be aggregated into a single rendering, which would otherwise blur temporal cues. 
\begingroup
\setlength{\abovedisplayskip}{3pt}
\setlength{\belowdisplayskip}{3pt}
\setlength{\abovedisplayshortskip}{3pt}
\setlength{\belowdisplayshortskip}{3pt}
\begin{equation}
\label{eq:surfel_score}
    \text{score}(s, t) =
    \frac{\langle \mathbf{n}_s, \bar{\mathbf{v}}_w \rangle}{1 + d_s}
    \cdot \ln(e + m_s)
    \cdot
    \left[
    \lambda_{\min}
    +
    \left(1 - \lambda_{\min}\right)
    2^{-\frac{T - t}{H}}
    \right],
\end{equation}
\endgroup
As shown in ~\cref{eq:surfel_score}, each rendered surfel at timestep $t$ is assigned a score based on geometric visibility, task relevance, and temporal recency. The geometric component depends on the depth $d_s$, normal $\mathbf{n}_s$, and its alignment with the average future wrist-camera viewing direction $\bar{\mathbf{v}}_w$. The task-relevance component is encoded by the \textit{is\_manipulated} flag $m_s$, while the temporal decay term assigns higher weights to more recently observed surfels. $T$ denotes  the latest timestep, $H$ denotes the temporal half-life, and $\lambda_{\min}$ denotes  the minimum temporal decay factor. This prioritizes observations that are both geometrically informative and closely related to the manipulation process. To further avoid redundancy, we apply non-maximum suppression (NMS) over the candidate observations. It mitigates oversampling of repeatedly visited regions, such as cases where the wrist camera hovers above the manipulated object during a pick operation, and promotes broader scene coverage among the selected top-$K$ timestep observations $o_g^1, o_g^2, \ldots, o_g^K$. The overall algorithm is listed in~\cref{alg:ctrl_mem}.

\vspace{-3mm}

%% file: sec/4_experiments.tex
\section{Experiments}
\vspace{-2mm}
In this section, we conduct experiments to evaluate Mem-World. We aim to answer the following
questions: (1) Can Mem-World generate consistent rollouts for long-horizon robot manipulation tasks with frequent end-effector occlusions and complex wrist camera motions? (2) Can Mem-World reliably evaluate different robot policies in imagination space on long-horizon tasks, faithfully reproducing their real-world performance? 
(3) Can Mem-World generate useful synthetic trajectories for improving policy performance on long-horizon manipulation tasks?

\textbf{Robot Setup and Dataset. }
Our experiments use a Franka Emika Panda robot arm equipped with a 2 DoF gripper. The platform includes one
wrist-view camera and two 
external third-view cameras that observe the workspace.
We use DROID~\citep{khazatsky2024droid} as the primary source for world-model training and replay-based evaluation. It 
contains 95,599 trajectories, providing diverse wrist camera motions.

\textbf{Training Details.} Our world model is built on top of Ctrl-World~\citep{guo2025ctrl-world}. To enable geometry-aware context conditioning, we sample 11K trajectories and construct training clips by replacing the original temporally adjacent context with geometry-related context retrieved by the training-free W-VMem. the $\lambda_{\min}$ in \cref{eq:surfel_score} is set to 0.1, the temporal half-life $H$ is set to $0.3\times T_{max}$, the We fine-tune only the temporal attention layers of the world model on these samples using 8 H100 GPUs with a total batch size of 32, which takes approximately 2 days. Considering the importance of the wrist view, we upweight the loss term associated with wrist-view predictions. 


\subsection{World Model Consistency Evaluation}

\begin{figure}[t]
    \centering
\includegraphics[width=1.0\linewidth]{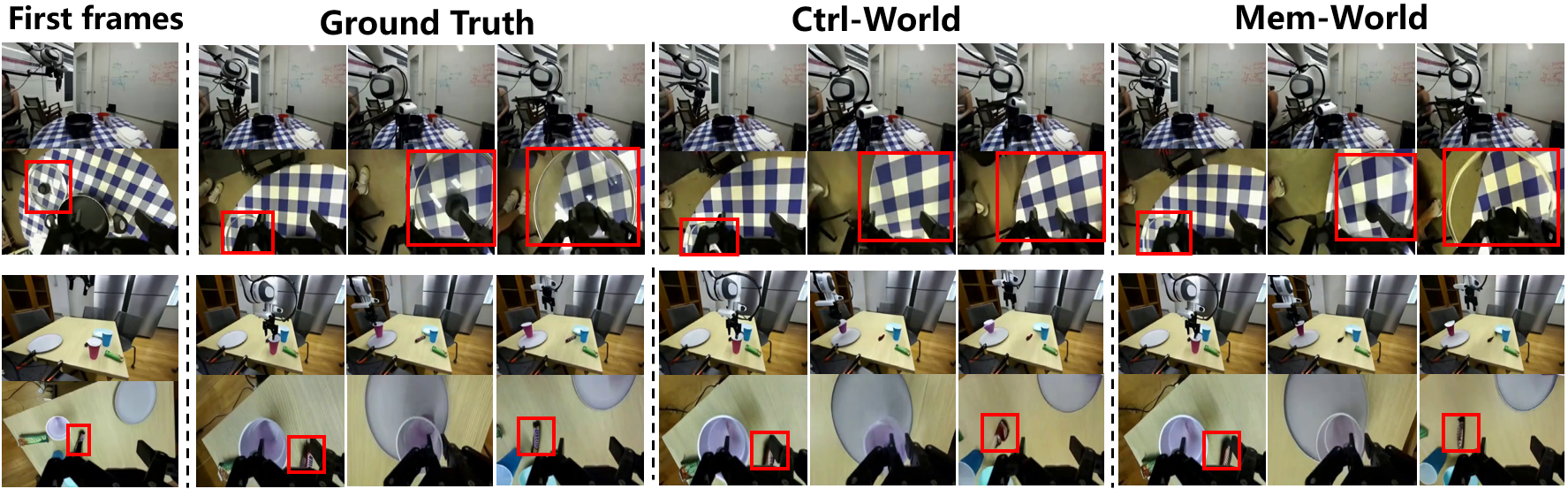}
    \caption{Qualitative results on long-horizon rollouts. Mem-World exhibits persistent and temporally consistent world modeling: the \textbf{lid} remains recoverable after transient occlusion by the gripper, and the \textbf{black snack} is consistently regenerated after back-and-forth camera motion.}
    \label{fig:compare_model}
\end{figure}

\textbf{Baselines and Evaluation Setup.} We compare Mem-World against two prior action-conditioned world models: Cosmos Predict 2.5~\citep{ali2025cosmos}, Ctrl-World. For Cosmos Predict 2.5, we finetune the pretrained checkpoint on single-view DROID data for third-person view action-conditioned world modeling. We do not extend this baseline to multi-view prediction due to the substantial computational cost.
To evaluate the temporal consistency, we carefully curate 34 trajectories from the DROID dataset for replay-based evaluation. Each trajectory satisfies at least one of the following conditions: (1) the manipulated object is temporarily occluded by the gripper in the wrist view and is later observed again, or (2) the wrist camera undergoes large back-and-forth motion. During rollouts, the world model autoregressively predicts future frames for 20 steps, producing 20-second trajectories. For evaluation metrics, we use PSNR~\citep{hore2010psnr}, SSIM~\citep{wang2004ssim}, and LPIPS~\citep{zhang2018lpips} to compare the predicted video with the ground-truth across all corresponding frames. In addition, we use SAM3~\citep{carion2025sam3} to segment the manipulated objects from both the predicted and ground-truth videos, and compute DINOv2~\citep{oquab2023dinov2} feature similarity to evaluate object-level consistency.  

\vspace{-8pt}
\begin{table}[h]
\centering
\caption{Quantitative world model consistency evaluation results. 
}
\resizebox{0.8\linewidth}{!}{
\begin{tabular}{c|c|cc|cc}
\toprule
\multirow{2}{*}{\shortstack{Evaluated \\ Camera}} 
& \multirow{2}{*}{Method} 
& \multicolumn{2}{c|}{Computation-based} 
& \multicolumn{2}{c}{Model-based} \\ 
\cmidrule(lr){3-4} \cmidrule(lr){5-6}
&  & PSNR $\uparrow$ & SSIM $\uparrow$ & LPIPS $\downarrow$ & Obj. Con. $\uparrow$  \\ 
\midrule

\multirow{3}{*}{\shortstack{Third-view \\ Camera}} 
& Cosmos Predict 2.5  & 22.80  & 0.819  & 0.089  & 0.579  \\ 
& Ctrl-World          & 23.17 & 0.828 & 0.076 & 0.573  \\
& Mem-World (Ours)              & \textbf{25.30} & \textbf{0.878} & \textbf{0.054} & \textbf{0.619}  \\

\midrule 

\multirow{2}{*}{\shortstack{Wrist-view \\ Camera}} 
& Ctrl-World          & 17.34 & 0.623 & 0.281 & 0.476  \\
& Mem-World (Ours)    & \textbf{19.21} & \textbf{0.691} & \textbf{0.236} & \textbf{0.524} \\

\bottomrule
\end{tabular}
}
\vspace{-2pt}
\label{tab:quant_results}
\end{table}

\textbf{Experiments Results.} 
As shown in~\cref{tab:quant_results} and~\cref{fig:compare_model}, Mem-World outperforms prior world models across computation-based and model-based metrics, indicating improved visual fidelity and temporal consistency in long-horizon rollouts. The gains are particularly significant on the wrist-view camera, which is more challenging due to rapid egocentric motion and frequent end-effector occlusions. We futher provide the time-cost analysis of W-VMem and addition replay-based evaluation on randomly sampled 150 trajectories to show the generalization of Mem-World in the supplementary materials.

\begin{wraptable}{r}{0.5\linewidth}
\vspace{-4pt}
\centering
\caption{Ablations on memory design. 
}
\vspace{-4pt}
\label{tab:ablation}
\scriptsize
\setlength{\tabcolsep}{2pt}
\renewcommand{\arraystretch}{0.82}
\resizebox{\linewidth}{!}{
\begin{tabular}{c|c|cc|cc}
\toprule
\multirow{2}{*}{\shortstack{Eval. \\ Cam.}} 
& \multirow{2}{*}{Method} 
& \multicolumn{2}{c|}{Comp.-based} 
& \multicolumn{2}{c}{Model-based} \\ 
\cmidrule(lr){3-4} \cmidrule(lr){5-6}
&  & PSNR $\uparrow$ & SSIM $\uparrow$ & LPIPS $\downarrow$ & Obj. Con. $\uparrow$ \\ 
\midrule

\multirow{3}{*}{\shortstack{Third \\ View}} 
& Short-term~\citep{wang2026interactivews} & 21.25 & 0.802 & 0.082 & 0.526 \\
& Stride~\citep{guo2025ctrl-world,jiang2025enerverse}     & 22.58 & 0.814 & 0.079 & 0.544 \\ 
& W-VMem (Ours)    & \textbf{24.78} & \textbf{0.869} & \textbf{0.062} & \textbf{0.597} \\

\midrule 

\multirow{3}{*}{\shortstack{Wrist \\ View}} 
& Short-term~\citep{wang2026interactivews} & 15.04 & 0.526 & 0.353 & 0.401 \\
& Stride~\citep{guo2025ctrl-world,jiang2025enerverse}     & 17.06 & 0.614 & 0.295 & 0.463 \\ 
& W-VMem (Ours)    & \textbf{18.97} & \textbf{0.680} & \textbf{0.248} & \textbf{0.502} \\

\bottomrule
\end{tabular}
}
\vspace{-2pt}
\end{wraptable}

\begin{figure}[t]
    \centering
    \includegraphics[width=0.92\linewidth]{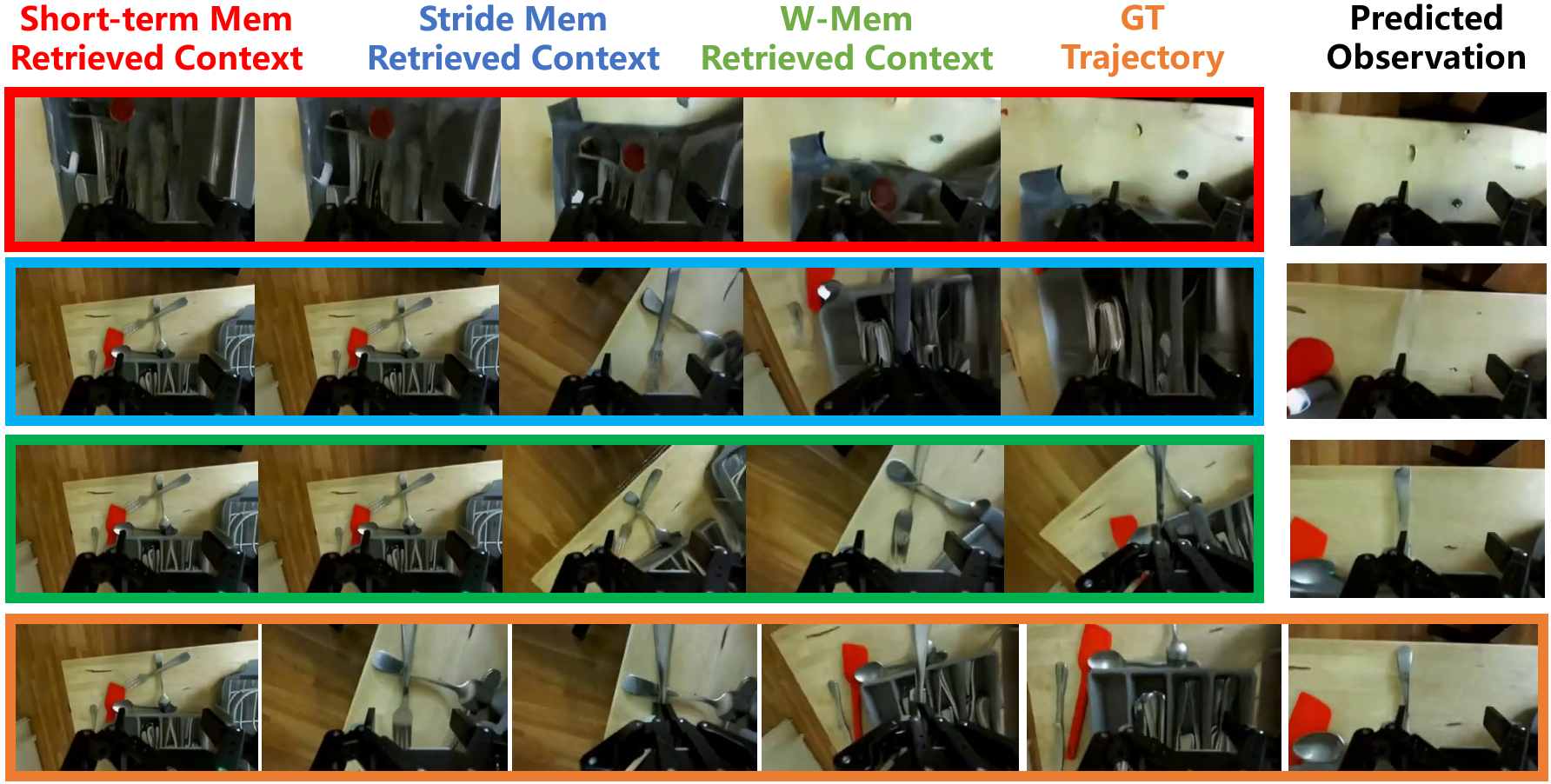}
    \caption{Ablation on memory retrieval strategies. By retrieving geometrically relevant historical frames, W-VMem helps the world model preserve consistent object appearances across future observations. In contrast, Short-term Mem and Stride Mem provide less informative context, leading to object (spoon) distortion or disappearance during rollout. }
    \label{fig:ablation}
\end{figure}

\textbf{Ablation Study of Memory Retrieval Method.}
We train a base model conditioned on randomly sampled context to fairly evaluate different memory retrieval methods. Here we compare our W-VMem, the stride memory adopted by Ctrl-World~\citep{guo2025ctrl-world} and EVAC~\citep{jiang2025enerverse}, and short-term memory adopted by Interactive World Simulator ~\cite{wang2026interactivews}, which selects $N$ adjacent frames as context on replay-based evaluations. As shown in \cref{tab:ablation} and \cref{fig:ablation}, W-VMem effectively retrieves relevant historical frames as context, allowing the world model to recover temporally occluded or previously observed scene content under back-and-forth camera motion, leading to more persistent predictions. Additional ablation study on the parameter of W-VMem can be found in supplementary materials.

\subsection{World Model for Policy Evaluation}

In this section, we evaluate whether Mem-World can serve as an effective simulator for evaluating the generalist robot policies, including $\pi_0$, $\pi_{0.5}$, and whether its predicted rollouts can accurately reflect their real-world performance. 

\textbf{Tasks Design. } 
We design five manipulation tasks with increasing temporal complexity. Two short-horizon tasks, \textit{pick cube to the plate} and \textit{wipe table}, require only a single pick-and-place operation or a unidirectional motion. Three long-horizon tasks, \textit{pick 1 object to 2 targets}, \textit{pick 2 objects to 1 target}, and \textit{pick 2 objects to 2 targets}, require two sequential subtasks with large loop-like wrist-camera motions. They progressively evaluate the world model's ability to preserve objects and targets during revisits, follow policy actions, and maintain long-term consistency across increasingly complex object-target configurations. More details can be found in the supplementary materials.

\begin{figure}[t]
\centering
\begin{minipage}[t]{0.38\linewidth}
\centering
\includegraphics[width=0.9\linewidth]{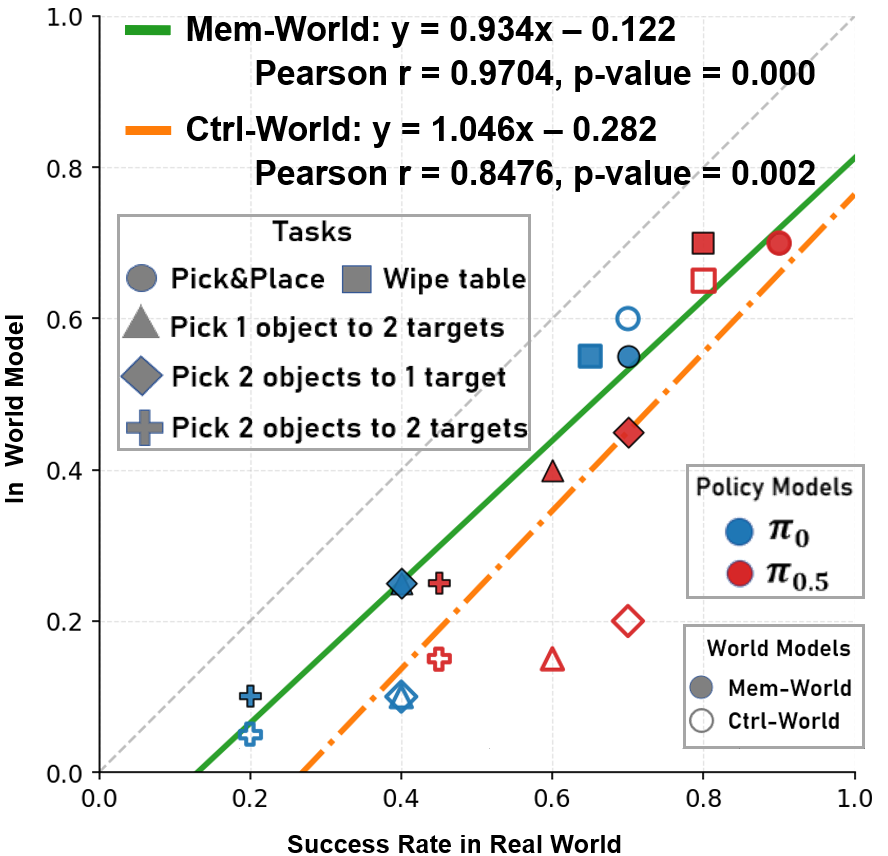}
\caption{Quantitative correlations between real-world and world-model.}
\label{fig:policy_eval_regression}
\end{minipage}
\hfill
\begin{minipage}[t]{0.55\linewidth}
\centering
\includegraphics[width=\linewidth]{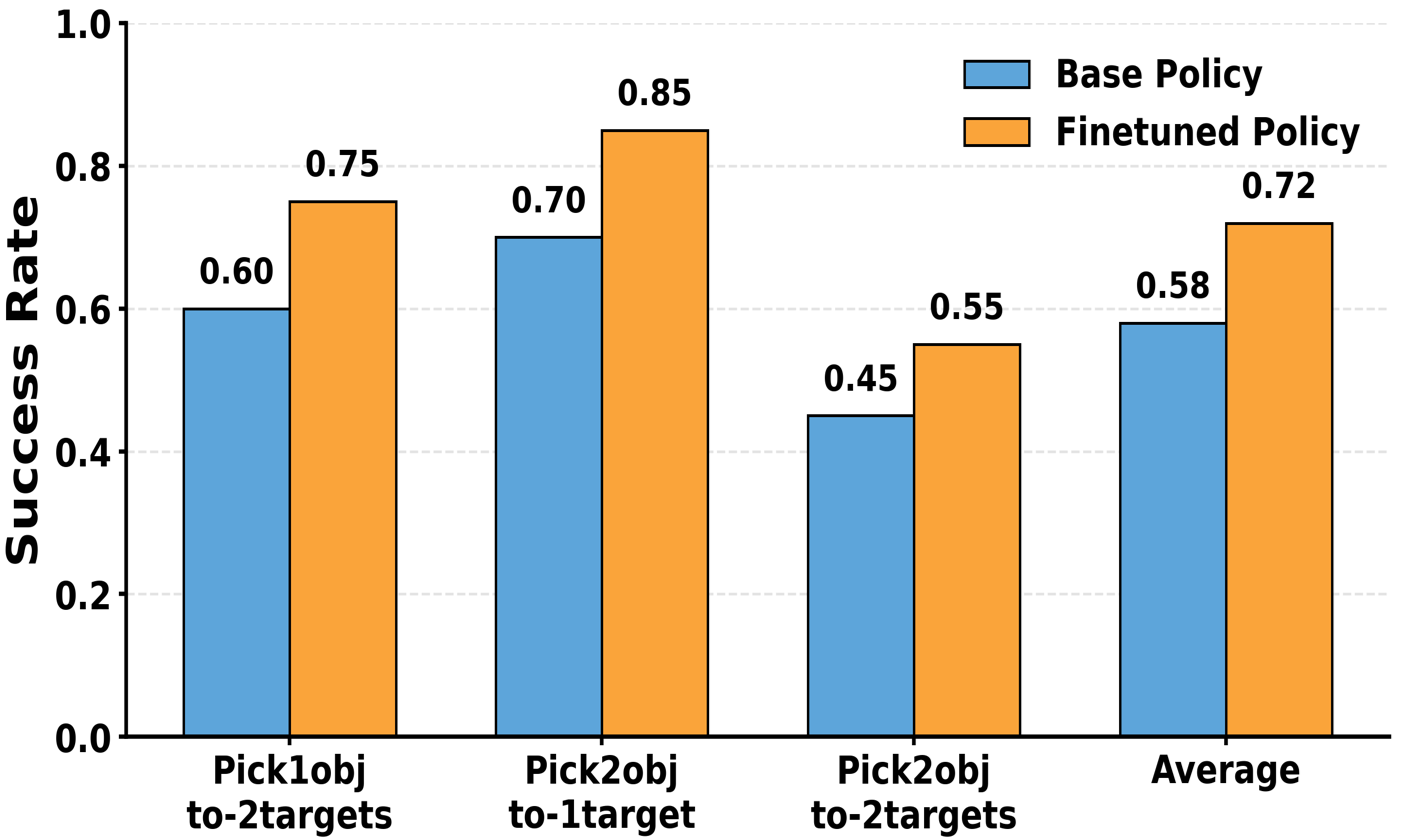}
\caption{Policy improvement. Post-training on synthetic data improves the average success rate to 72\%.}
\label{fig:policy_improve}
\end{minipage}
\end{figure}

\begin{figure}[t]
\centering
    \includegraphics[width=1.0\linewidth]{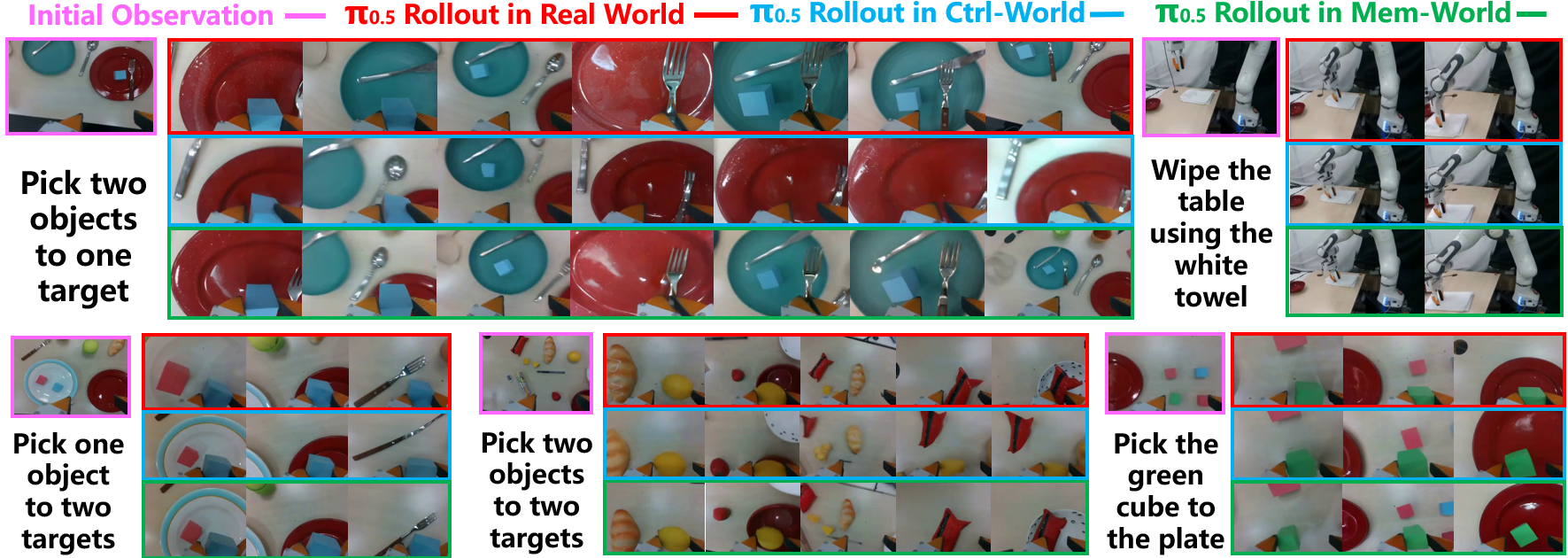}
\caption{Comparisons between $\pi_{0.5}$ rollouts in the real-world, Ctrl-World, and Mem-World.}
\label{fig:policy_eval}
\end{figure}

\textbf{Comparison Between Real-World and World Model Rollouts.}
We initialize real-world executions and world-model rollouts, including Ctrl-World and Mem-World, from the same initial observations and evaluate each policy under identical task conditions. Since our setup is out-of-distribution for both policy and world models due to the custom environment and different 2DoF gripper, we collect 50 episodes per task for post-training. We fine-tune $\pi_0$ and $\pi_{0.5}$ for 20K steps on 4 H100 GPUs, and fine-tune the world model for 5K steps on 4 H100 GPUs to adapt to the visual domain shift. Policy evaluation is then conducted using the post-trained models. 
As shown in \cref{fig:policy_eval_regression}, policy success rates in Mem-World are highly correlated with their real-world success rates across the five tasks ($r = 0.97$, $p < 0.001$). Although the absolute success rates in Mem-World are generally lower than those in the real world, the gap is relatively uniform across tasks, with no obvious outliers, resulting in a strong linear correspondence. In contrast, Ctrl-World shows a weaker correlation ($r = 0.85$, $p < 0.01$), mainly due to its pronounced performance drop on long-horizon tasks. This deviation suggests that the stride-memory design struggles to preserve scene consistency during long-horizon rollouts, leading to poorer alignment with the real world. As shown in \cref{fig:policy_eval}, in the \textit{pick 2 objects to 1 target} task, Ctrl-World severely distorts the fork, preventing subsequent grasp completion.

\vspace{-1mm}
\subsection{World Model for Policy Training}
\vspace{-1mm}
We now evaluate whether Mem-World can be used to generate synthetic post-training data for improving policy models. We use the $\pi_{0.5}$ as a base policy to rollout with Mem-World in initial observation with unseen object arrangements. We generate 200 trajectories per task and retain 20--30 successful trajectories based on human success annotation, ensuring high-quality task-completion behaviors.
we compare the post-trained $\pi_{0.5}$ policy before and after additional fine-tuning on Mem-World-generated successful trajectories. On the three long-horizon tasks, synthetic post-training improves the average real-world success rate from 58\% to 72\%, as shown in \cref{fig:policy_improve}.

\vspace{-1mm}


%% file: sec/5_conclusion.tex
\section{Conclusion}
\vspace{-1mm}
We introduce Mem-World, a memory-augmented multi-view action-conditioned world model for persistent long-horizon robot manipulation simulation. To address inconsistencies caused by end-effector occlusions and rapid wrist-camera motion during manipulation, W-VMem is introduced to retrieves the relevant past observations for future prediction. Experiments show that Mem-World produces more persistent rollout than prior works and widely adopted memory mechanisms. Moreover, its rollouts correlate well with real-world policy performance and can improve policy training when used as a data engine. These results highlight memory-augmented world modeling as a promising step toward reliable and physically grounded simulation for robot learning.

%% file: sec/6_limitation.tex
\textbf{Limitation.} While Mem-World demonstrates encouraging results, there remain several directions for future improvement. First, the rollout quality can be influenced by the informativeness of the initial wrist-view frame. In cases where this frame is affected by occlusion or a limited field of view, the model may have insufficient visual evidence for maintaining consistency in later predictions. A promising direction is to further enhance multi-view consistency to better leverage complementary observations from other viewpoints for more persistent rollout. Second, 
Mem-World does not explicitly enforce physical constraints.
For example, when grasping a spherical object, the world model may still predict a successful grasp even when the gripper fails to establish antipodal contacts. 
Future work may incorporate physics-aware signals, such as contact or force information, to further improve the physical fidelity of imagined rollouts.